\documentclass{article} % For LaTeX2e
\usepackage{iclr2026_conference,times}

\usepackage{amsmath,amsfonts,bm}

\def\eqref#1{equation~\ref{#1}}
\def\1{\bm{1}}

\DeclareMathAlphabet{\mathsfit}{\encodingdefault}{\sfdefault}{m}{sl}
\SetMathAlphabet{\mathsfit}{bold}{\encodingdefault}{\sfdefault}{bx}{n}

\DeclareMathOperator*{\argmin}{arg\,min}

\usepackage{graphicx}
\usepackage{hyperref}
\usepackage{url}
\usepackage{algorithm}
\usepackage[noend]{algorithmic}

\usepackage{multirow}
\usepackage{placeins}

\usepackage{amsmath}
\usepackage{amsfonts}
\usepackage{amsthm}
\usepackage{cleveref}
\usepackage{nicefrac}
\usepackage{booktabs}

\newcommand{\ouracronym}{{GP-Surrogate}}    % acronym what we will use for our method

\newcommand{\changed}[1]{#1}

\title{Scalable Decision Focused Learning via Online Trainable Surrogates}

\author{Gaetano Signorelli and Michele Lombardi \\
\texttt{University of Bologna} \\
\texttt{\{gaetano.signorelli2, michele.lombardi2\}@unibo.it}}

\iclrfinalcopy % Uncomment for camera-ready version, but NOT for submission.
\begin{document}

\maketitle

\begin{abstract}

Decision support systems often rely on solving complex optimization problems that may require to estimate uncertain parameters beforehand.
Recent studies have shown how using traditionally trained estimators for this task can lead to suboptimal solutions.
Using the actual decision cost as a loss function (called Decision Focused Learning) can address this issue, but with a severe loss of scalability at training time.
To address this issue, we propose an acceleration method based on replacing costly loss function evaluations with an efficient surrogate.
Unlike previously defined surrogates, our approach relies on unbiased estimators -- reducing the risk of spurious local optima -- and can provide information on its local confidence -- allowing one to switch to a fallback method when needed.
Furthermore, the surrogate is designed for a black-box setting, which enables compensating for simplifications in the optimization model and accounting for recourse actions during cost computation.
In our results, the method reduces costly inner solver calls, with a solution quality comparable to other state-of-the-art techniques.

\end{abstract}

\section{Introduction}

Many real-world decision support systems, in domains such as logistics or production planning, %or supply chain management,
rely on the solution of constrained optimization problems with parameters that are estimated via Machine Learning (ML) predictors.
Literature from the last decade has showed how this approach, sometimes referred to as Prediction Focused Learning (PFL), can lead to poor decision quality due to a misalignment between the training objective (usually likelihood maximization) and the actual decision cost.
Decision Focused Learning (DFL) \citep{amos2017optnet,elmachtoub2022smart} was then introduced to correct for this issue by training predictors for minimal decision regret.

While remarkable progress in the field has been made \citep{mandi2024decision}, we argue that, \changed{based on our experience with industrial optimization use cases}, three issues still prevent DFL from finding widespread practical application.
First, \changed{while DFL methods are very efficient at inference time,} their \emph{training scalability is often severely limited}, since the problems encountered in decision support are frequently difficult (NP-hard or worse) and most DFL approaches require frequent solver calls and cost evaluations.
Second, many DFL methods make \emph{restrictive assumptions} on the decision problem (e.g. linear cost function, no parameters in the constraints);
\changed{in addition to limiting applicability, it has been shown \citep{hu2022fast,elmachtoub2023estimate} that such assumptions also cause the DFL advantage to vanish if the parameters expectations can be accurately estimated (\cref{appendix:dfl_classical})}.
Third, several DFL methods require \emph{explicit knowledge of the problem structure or the solver state}, which in a practical setting would require costly refactoring of the existing tools, or even a radical change of the solution technology.
Solving these issues would allow one to use DFL for improving the effectiveness and robustness of \emph{any real-world decision support tool}, while maintaining scalability.

% Many real-world decision support systems, in domains such as logistics or production planning, %or supply chain management,
% rely at their core on the solution of a constrained optimization problem.
% The problems arising in these domains are often so complex (NP-hard or worse) that their efficient solution has long represented a major bottleneck and a major focus for research attention.

% However, in a practical setting, suboptimal decisions might arise not just due to poor exploration of the solution space, but also due to \emph{the context in which these systems are employed}.
% Notably, it has been shown in recent years \citep{amos2017optnet,elmachtoub2022smart}
% that the common practice of replacing uncertain parameters with estimates obtained via a predictor can lead to poor quality in decisions even when the core problem is solved to optimality.
% This is due to a misalignment between the training objective (usually likelihood maximization) and the actual decision cost.
% This observation has lead to Decision Focused Learning (DFL) \citep{mandi2024decision}, which attempts to correct for this issue by training predictors for minimal decision regret.
% As a major downside, these methods require solving multiple, computationally expensive, optimization problems during training, greatly diminishing their practical appeal.

We aim at making a significant step toward addressing these limitations, by relying on a carefully designed, efficient, surrogate to replace most solver calls at training time.
\changed{Our surrogate is suitable for a black-box setting, where no restrictive assumption is made on regret computation and no access to the solver state is needed.
Compared to the relevant state of the art: 1) our surrogate is an asymptotically unbiased regret estimator, i.e. with no irreducible approximation error;
2) we use a principled mechanism (stochastic smoothing and importance sampling) to address 0-gradients often occurring in DFL settings;
3) we include uncertainty quantification via a confidence level, used to decide when to dynamically update the surrogate based on samples generated by a fallback method.}
% and hence balance exploration and  exploitation at training time.}
% This choice enables compensating for a second source of misalignment often present in real-world decision support systems, i.e. approximations in the decision model.
% In fact, a decision vector that is nominally optimal and feasible may fail to be so in practice, due to compromises and simplifications made when formulating the problem model itself, for example to improve scalability in decision-making under uncertainty.

% We propose a method suitable for general black-box functions, with no assumptions on the underlining optimization problem, which is an advantage over the numerous state-of-the-art models requiring a problem relaxation under theoretical assumptions (e.g., a MILP problem).
Only a few of the existing DFL methods can be applied to achieve similar goals, \changed{a representative set of which is used as a baseline in our empirical evaluation}.
\changed{We design our experiments to assess the scalability and effectiveness of our method in a controlled setting, by comparing} DFL and PFL on extended versions of standard benchmarks in the current literature.
\changed{We emphasize problems with recourse actions and/or non-linearities, since they represent the settings where the benefits of DFL over PFL are robust even when accurate predictions can be obtained}.
\changed{We allow for} scaling the problem complexity, to assess how the evaluated approaches behave on problems of different size (in terms of number of variables or parameters).
In our results, our surrogate significantly reduces both the training runtime and the number of solver calls and cost evaluations.
We also show that this acceleration, unlike previous attempts in the literature, does not adversely affect the decision quality, which remains comparable to other state-of-the-art techniques.

% \begin{itemize}
%     \item Main selling points
%     \begin{itemize}
%         \item Generality (support for black-box cost functions)
%         \item Significant reduction in solver and cost evaluation calls
%     \end{itemize}
%     \item Problem classes:
%     \begin{itemize}
%         \item Main focus: stochastic optimization problems
%         \item Also: problem where cost evaluation requires simulation
%         \item Also: any classical DFL scenario
%     \end{itemize}
%     \item Method
%     \begin{itemize}
%         \item Using a surrogate
%         \item The surrogate is unbiased
%         \item Solving the 0-gradient problem
%         \begin{itemize}
%             \item The surrogate itself
%             \item Stochastic smoothing
%         \end{itemize}
%         \item Uncertainty quantification to determine when to use (and consequently when to update the surrogate)
%     \end{itemize}
% \end{itemize}

\section{Related Work}

% In this section, we provide an overview of related literature, focusing on DFL techniques and methods for black-box optimization.
% For a more comprehensive overview of DFL approaches, we refer the reader to \citet{mandi2024decision}.

% \paragraph{Decision Focused Learning}

In the context of DFL problems where parameters are predicted by a machine learning model, initial works focused on implicit differentiation of the KKT conditions for optimality.
In particular, \citet{amos2017optnet} handled convex quadratic programming, while \citet{agrawal2019differentiable} extended it to conic programs. However, these initial methods were unsuitable for combinatorial problems, characterized by piecewise constant loss functions and uninformative gradients.
Subsequent studies such as \citet{wilder2019melding,mandi2020interior} addressed MILP problems, by proposing to smooth the loss function through a regularization term (respectively L2 and log-barrier) computed over the decision variables.
% : the Euclidean norm of the decision variables in the first case, and the log-barrier term in the second one.
Other approaches introduced surrogate losses to overcome the zero-gradient problem. \citet{elmachtoub2022smart} formalized the SPO+ loss as a regret upper bound, \citet{mulamba2020contrastive} proposed the noise-contrastive estimation, \citet{mandi2022decision} turned the problem into learning to rank on a pool of solutions. Many of these works also adopted a LP relaxation to speed up the training, which can however adversely affect the final solution quality, and it requires formulating the decision problem as a MILP.
A surrogate loss, based on directional gradients, is also proposed in \citet{huang2024decision} and proved to provide unbiased gradient estimates, which allows it to outperform earlier approaches for strongly misspecified ML models.
Unlike our approach, however, this method is restricted to problems with linear cost functions and is actually more computationally expensive at training time, due to the use of a zeroth-order gradient approximation.
% This recent work, similarly to our method, makes use of an unbiased estimator (assuming enough data is available), but it is still limited by a high computational demand and the linearity assumption of the optimization problem. Moreover, our approach finds a different target in addressing misspecifications related to the black-box optimization problem itself.
Finally, the approaches mentioned so far, with the exception of the KKT-based solutions, do not allow for predicted parameters appearing in the problem constraints.
Attempts to cover the latter case include \citet{paulus2021comboptnet,hu2023branch,hu2023predict+,hu2023two}, which require access to the problem formulation and either dedicated solvers or access to the solver internal state.

Only a few DFL approaches target the setting considered in this paper, where no assumption is made on the problem structure and training time scalability is emphasized; this is typically done by replacing the decision cost loss with fast-to-evaluate, differentiable, and learnable surrogates.
The first studies in this class include \citet{chung2022decision,lawless2022note}, which used simple loss approximations with poor results.
Recent advances are represented by the convex learnable surrogates proposed by \citet{shah2022decision,shah2024leaving}, and LANCER \citep{zharmagambetov2023landscape}, which employs a trainable neural network surrogate, fine-tuned at training time similarly to actor-critic approaches in reinforcement learning.
These methods are closest to the one we propose, and differ mainly for their use of biased estimators and the lack of local confidence estimation.

% A last class of methods has focused on using fast-to-evaluate surrogates to reduce the number of solver calls and achieve differentiability in general cases, with no assumptions on the problem structure.
% These works allow to address general settings, while limiting the number of solver calls, hence addressing the two most relevant limitations of the previous methods.
% However, they often lead to suboptimal solutions in classical DFL problems.
% Our approach belongs to this last category, as it aims at accelerating the training runtime, while being applicable to general settings and keeping the performance on par with that of state-of-the-art methods.

% \paragraph{Bayesian Optimization and Reinforcement Learning}

DFL problems can also be formulated by treating the decision problem similarly to a reinforcement learning environment, as suggested by \citet{silvestri2022unify}.
% This perspective enables addressing these problems using RL and black-box optimization techniques. 
A notable example is the Score Function Gradient Estimation (SFGE) method by \citet{silvestri2023score}, inspired by \citet{donti2017task,poganvcic2019differentiation,berthet2020learning,mohamed2020monte} and \citet{niepert2021implicit}, which combines stochastic smoothing and policy gradient methods.
% Our work relies on the same stochastic smoothing combined with a contextual Bayesian optimization strategy to approximate regret functions.
While we rely on Gaussian processes like \citet{char2019offline}, we use separate models and exploit contextual information via sample sharing, which simplifies the learning process and avoids length-scale and kernel issues.

\section{Problem Formulation}

We consider a generalization of a DFL setting, where the parameters $y$ of an optimization problem (e.g. demands) are not known at decision time, but can be estimated based on contextual information $x$ (e.g. hour of the day).
Formally, let $X$ and $Y$ be random variables with support $D_x$ and $D_y$, representing respectively contextual information and the uncertain parameters, and correlated according to their joint distribution $P(X, Y)$.
At decision-making time, the problem parameters are estimated via a predictive model $h_\theta: D_x \rightarrow D_y$, with parameter vector $\theta$.
Based on the estimator output $\hat{y}$, we compute a decision vector $z^*$ by solving a constrained optimization problem:
\begin{equation}
    z^*(\hat{y}) = \argmin_z \left\{ f(\hat{y}, z) \mid z \in C(\hat{y}) \right\}
    \label{eqn:opt_problem}
\end{equation}
where $f : D_y \times D_z \rightarrow \mathbb{R}$ is the problem cost function and $C: D_y \rightarrow 2^{D_z}$ is a constraint function that denotes the feasible space.
We treat $z^*$ as a function, assuming tie-breaking is used when multiple optimal solutions exist.
Once the decisions are executed, their quality is determined by means of a second ``true'' cost function $g : D_y \times D_z \rightarrow \mathbb{R}$.
Specifically, $g(y, z)$ represents the cost incurred by the solution $z$, under a realization $y$ sampled from $P(Y \mid x)$.

The use of a distinct function $g$ for decision quality evaluation distinguishes our setup from those typically used in DFL, and enables compensating for \emph{misspecified decision problem models} -- as opposed to misspecified predictors as in \citet{huang2024decision}.
These can stem from treating uncertain parameters (e.g. travel times or demands) as deterministic, from disregarded minor constraints, or from approximated non-linearities -- all common techniques to ensure scalability in real-world applications.
This choice also allows us to deal with estimated parameters in the problem constraints in \cref{eqn:opt_problem}, assuming that infeasible solutions can be repaired at an additional cost.

% recourse actions in two-stage stochastic programming, simulation-based cost functions, or disregarded non-linearities in the optimization problem formulation.

\changed{We wish to train $h$ for minimal expected decision regret, approximated via a sample average:}
\begin{equation} \label{eq:sample-based-training}
    \argmin_\theta \frac{1}{m} \sum_{i=1}^m 
        r(y_i, \hat{y}_i)
\end{equation}
where $\{x_i, y_i\}_{i=1}^{m} \sim P(X, Y)$ is the training data, $\hat{y}_i = h_\theta(x_i)$ and $r(y_i, \hat{y}_i) = g(y_i, z^*(\hat{y}_i)) - g(y_i, z^*(y_i))$.
% \begin{equation} \label{eq:training problem}
%     \argmin_\theta \mathrm{E}_{x, y \sim P(X, Y)} \left[
%         r(y, \hat{y})
%     \right]
% \end{equation}
% where $\hat{y} = h_\theta(x)$ and $r(y, \hat{y}) = g(y, z^*(\hat{y})) - g(y, z^*(y))$.
% In practice, the expectation is approximated by an average over a training sample $\{x_i, y_i\}_{i=1}^{m}$, thus leading to:
% \begin{equation} \label{eq:sample-based-training}
%     \argmin_\theta \frac{1}{m} \sum_{i=1}^m 
%         r(y_i, \hat{y}_i)
% \end{equation}
% where $\hat{y}_i = h_\theta(x_i)$.
% The predictive model is trained to minimize the true decision cost, compensating for both uncertainty in the parameters and modeling simplifications.
% As a notable use case, in real-world decision support tools it is common to treat uncertain parameters as deterministic in the optimization problem, for sake of scalability.
% In this setup, training the predictor as in \cref{eq:sample-based-training} allows to account for uncertainty -- by means of the true cost $g$ -- while retaining the scalability of the deterministic solver.
% In a real-world use case, this can be extremely appealing, though it has been recently shown that a similar formulation might incur a structural loss of optimality due to lack of expressivity in some cases \citep{schutte2025sufficient}.
%
Solving \cref{eq:sample-based-training} via first-order methods, as typically done with neural networks predictors, requires computing the gradient of the regret function, i.e.:
\begin{equation}
    \frac{\partial }{\partial \theta} r(y_i, \hat{y}_i) = \frac{\partial g}{\partial z^*} \frac{\partial z^*}{\partial \hat{y}_i} \frac{\partial \hat{y}_i}{\partial \theta}
\end{equation} 
When the optimization problem is linear or combinatorial, or the $g$ function is piecewise-constant, the gradient might be undefined or null on a large part of the predicted parameter space.
Furthermore, evaluating the regret function requires computing $z^*$ and $g$ once per example and per training epoch, which can be prohibitively expensive, when the optimization problem is NP-hard, or the true cost function is based on optimization or simulation.

% Ideally, a training approach for \cref{eq:sample-based-training} should provide a gradient approximation that allows a trade off between accuracy -- to avoid spurious local optima -- and usability -- to avoid getting stuck on flat regions of the parameter space.
% Additionally, such as approach should limit as much as possible the number of calls to both $z^*$ and $g$.

\section{Methodology}

We now introduce our method, whose main goal is accelerating the training problem of \cref{eq:sample-based-training}.
Similarly to \citet{shah2022decision,shah2024leaving}, we reduce the runtime by replacing, for every training example, the computationally heavy loss function $r(y_i, \hat{y})$ with a faster, trainable, surrogate loss $\tilde{r}_i(\hat{y})$.
Formally, for a given realization $y_i$, associated to the $i$-th training example, the surrogate model is a function $\tilde{r}_i : D_y \rightarrow \mathbb{R}$ mapping a prediction vector $\hat{y}$ into a corresponding loss value.

We identify three desirable properties for such a function.
First, the surrogate should be \emph{differentiable and have informative gradients everywhere}, to support gradient-descent optimization.
Second, $\tilde{r}_i$ should be capable of providing \emph{unbiased gradient estimation} to ensure that, if enough calibration data is available, the local optima of the regret loss are preserved.
% for convergence to real optima points for both stochastic and deterministic problems.
The surrogate losses from \citet{chung2022decision,lawless2022note,shah2022decision,shah2024leaving} do not satisfy this criterion, running the risk of getting trapped in spurious local minima.
Third, $\tilde{r}_i$ should provide \emph{confidence information} for online refinement; 
% point-wise information on its uncertainty.
namely, one should be able to determine when the surrogate is reliable, and when instead a fallback method based on direct evaluation of $z^*$ and $g$ should be employed.
% last point enhance reliability: it allows to alternate between exploration and exploitation by calling a fallback method whenever the surrogate is not confident.

We propose using Gaussian Processes (GP) with Radial Basis Function (RBF) kernels for the surrogate, since they satisfy almost all the desired properties.
In particular, GPs with RBFs are fully differentiable and can support efficient evaluation.
Moreover, GPs are (asymptotically) unbiased estimators: given enough samples, they can approximate any function with arbitrarily high precision -- though they are biased towards zero-centered predictions with scarce data.
Finally, GPs have distributional output and naturally provide confidence information.
% and, in addition, they are simple and easy to train models, especially as, given enough points, they can approximate any function using few learnable parameters. For this reason they represent convenient unbiased estimators. Moreover, a GP is a statistical model that predicts normal distributions, thus providing the requested point-wise confidence.

\begin{figure}[tbh]
    \centering
    \includegraphics[width=0.6\textwidth]{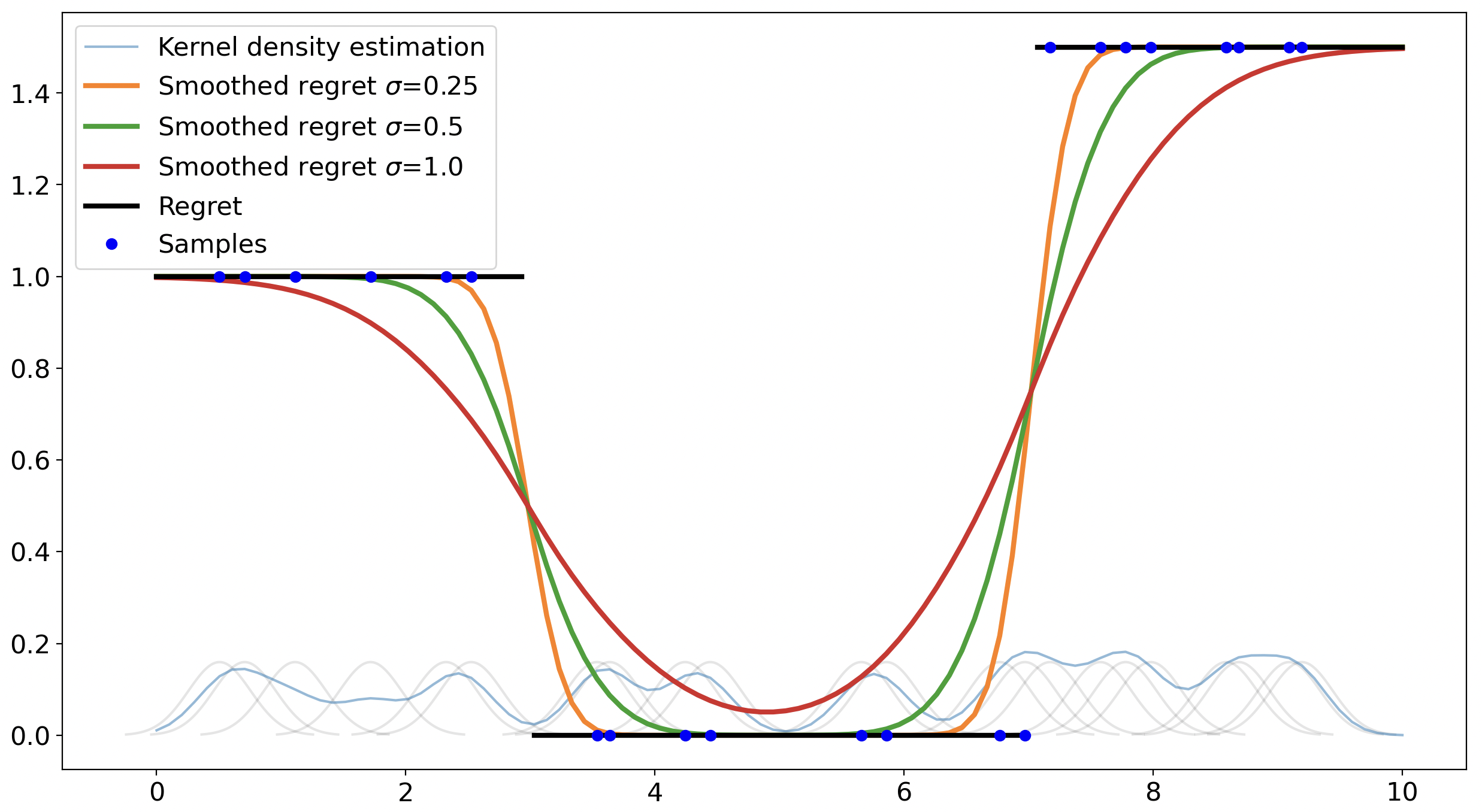}
    \caption{Illustration of stochastic smoothing with varying $\sigma$: larger values cause smoother functions with respect to the original regret loss. Smoothed functions are computed using importance sampling over a set of points sampled from Normal distributions.}
    \label{fig:stochastic_smoothing}
\end{figure}

\paragraph{Stochastic smoothing}
The only property that GPs lack is tied to their nature as unbiased estimators.
On the one hand, their are capable of accurately approximating the regret function; on the other hand, they risk inheriting some of its undesirable traits, such as 0 (or near-0) gradients on large swathes of the input space.
% the 0-gradient problem typical of piecewise-constant loss functions.
% This is a common case in DFL settings, where regret is often constant under small adjustments of the predictor output.
One possible solution to this issue is to apply stochastic smoothing to the loss function -- somewhat similarly to \citet{silvestri2023score}.
Specifically, our surrogate approximates a \emph{smoothed} version of $r_i$, here called $\bar{r}_i$, referred to as $\bar{r}_i: D_y \rightarrow \mathbb{R}$ and defined as:
% \begin{equation}
%     \bar{r}_i: D_y \rightarrow \mathbb{R}
% \end{equation}
% where:
\begin{equation} \label{eq:smoothed r}
    \bar{r}_i(\hat{y}) = \mathrm{E}_{\hat{y}' \sim \mathcal{N}(\hat{y}, \sigma)} \left[
        r_i(\hat{y}')
    \right]
\end{equation}
The original loss value is replaced with its expectation under random, Normally distributed, perturbations of its input.
The standard deviation $\sigma$ is a controllable parameter in the method and represents the degree of smoothing.
Small values of $\sigma$ result in a more accurate regret approximation -- at the cost of possibly weaker gradients -- while higher $\sigma$ values can provide more informative gradients -- but may result in altered local optima.
The effect of smoothing is depicted in \cref{fig:stochastic_smoothing}.

% Importantly, $\bar{r}_i$ smooths the real loss function with possible alterations of local optima depending on $\sigma$ \citep{silvestri2023score}.

In principle, $\bar{r}_i$ can be computed via a simple Monte Carlo approach.
In practice, this is highly inefficient, because many samples would be needed for each input $\hat{y}$, and evaluating $r_i(\hat{y}')$ for each sample requires computing both $z^*$ and $g$.
% it requires to sample multiple values of $\hat{y}$ with multiple computations of $r_i(\hat{y}')$ to approximate the expected value at each point.
We overcome this limitation by relying on importance sampling to perform multiple computations of $\bar{r}_i(\hat{y})$ based on the same set of observations.
% To overcome this computational limitation, we adopt the importance sampling technique as a Monte Carlo estimation for the expected value.
Formally, let $\{\hat{y}'_k \}_{k=1}^n$ be a set of predictions for which the value of $r_i(\hat{y}'_k)$ is known.
We assume each of them has been sampled according to a distinct process, and refer to $\{q_k\}_{k=1}^n$ as the corresponding probabilities. These samples can be associated to an aggregated probability density function $q$, defined as a kernel density estimator, and used to define the importance weight function $w(\hat{y}, \hat{y}')$:
\begin{align}
    q(\hat{y}) &= \frac{1}{n} \sum_{k=1}^n q_k(\hat{y})
    &
    w(\hat{y}, \hat{y}') &= \frac{\phi(\hat{y}'; \hat{y}, \sigma)}{q(\hat{y}')}
\end{align}
%
% \begin{equation}
%     q(\hat{y}) &= \frac{1}{n} \sum_{k=1}^n q_k(\hat{y})
% \end{equation}
% %
% Then, we define the importance weight function $w$ as:
% %
% % \begin{equation}
% %     w(\hat{y}, \hat{y}') = \frac{P(\tilde{y}=\hat{y}')}{q(\hat{y}')}, \quad \tilde{y} \sim \mathcal{N}(\hat{y}, \sigma)
% % \end{equation}
% \begin{equation}
%     w(\hat{y}, \hat{y}') = \frac{\phi(\hat{y}'; \hat{y}, \sigma)}{q(\hat{y}')}
% \end{equation}
%
where $\phi(\cdot; \hat{y}, \sigma)$ is the density for a Normal distribution centered on $\hat{y}$ and having standard deviation $\sigma$.
% where $\sigma$ is a parameter controlling the smoothing factor. Then, given a set of points $Y' \subset Y$, it holds:
Then we have:
% \begin{equation} \label{eq:importance sampling}
%     \bar{r}_i(\hat{y}) \approx \sum_{\hat{y}' \in Y'} \frac{r_i(\hat{y}_i') \cdot w(\hat{y}, \hat{y}')}{\sum_{\hat{y}' \in Y'} w(\hat{y}, \hat{y}')}
% \end{equation}
\begin{equation} \label{eq:importance sampling}
    \bar{r}_i(\hat{y}) \approx \sum_{k = 1}^n \frac{w(\hat{y}, \hat{y}'_k)}{\sum_{h=1}^n w(\hat{y}, \hat{y}'_h)} r_i(\hat{y}_k')
\end{equation}
In practice, whenever a new prediction $\hat{y}_k'$ is evaluated during the training process, we store both its regret value and the probability according to which it was sampled.
\changed{When training our GPs, we use as target the \emph{smoothed} regret estimated via \cref{eq:importance sampling}}.
% finally, we use the smoothed values as targets when training our GPs.
% Practically, we use the GP training points, for which the loss value as already been computed, as $Y'$, and the distributions they have been sampled from as $Q$. 
This approach allows to increase the sampling efficiency and decrease the variance associated to the computation of $\bar{r}_i$; at the same time it adds the flexibility to control the smoothing level by manipulating the sampling distributions.
It is worth noting that, when a limited number of sampling points is available, the natural smoothing from the GPs combines with stochastic smoothing, leading to very regular landscapes for the surrogate loss.
In \cref{fig:stochastic_smoothing} we show an example of stochastic smoothing via importance sampling computed on a set of one-dimensional points sampled from Normal distributions.

\changed{In \Cref{appendix:estimation_bias} we discuss how stochastic smoothing affects our surrogate bias, and we show that our approach still has no irreducible error provide that abundant data is available and $\sigma$ is small enough.
In practice, higher $\sigma$ values and few solver calls will be desirable, but the use of an expressive surrogate can decrease the risk of getting trapped in poor quality local optima.}

% We point out that this solution for the 0-gradient problem is particularly effective in combination with a GP surrogate, if only few training samples are available. In fact, stochastic smoothing is combined with the natural smoothing of a GP, due to its weighted interpolation mechanism;

\paragraph{Sample sharing}

Using a distinct surrogate for each training example permits unbiased regret estimation, at the same time limiting the number of input dimensions for the GPs.
However, this also prevents information sharing among surrogates.
With the aim to further reduce the number of $r_i$ function evaluations, we devised an optional technique to enable sample sharing between different surrogates, if the corresponding regret landscapes are similar.
% As a further speed-up technique, we include in our method information sharing between similar GPs, so to reduce the number of calls to the real $r_i$ function.
% To measure similarity, we only consider the target space:
Specifically, at the beginning of the training process, we perform Latin Hypercube Sampling (LHS) to collect a number points $\{\hat{y}_k \}_{k=1}^m$ in the prediction space $D_y$.
We then associate each training example with a vector $v$ containing the regret value computed for each such collected point, i.e. $v_i = \{r_i(\hat{y}_k)\}_{k=1}^m$.
It can be seen that two samples $i$ and $j$ are associated to the same regret landscape iff:
\begin{equation}
    \lim_{m \to \infty} \|v_{i} - v_{j}\|_2 = 0
\end{equation}
i.e., if the Euclidean distance between the corresponding vectors converges to 0, as the number of sampled points grows.
Accordingly, we measure the similarity between the regret landscapes for two samples $i$ and $j$ in terms of such distance.
%
% we apply the Latin Hypercube Sampling (LHS) to sample $Y_k = \{\hat{y}_1, \hat{y}_2, \dots, \hat{y}_k\} \subset Y$ evenly spaced points across the dimensions, and build $m$ vectors of the form $v \in \mathbb{R}^k \mid v = r_i(Y_k)$; then we build a distance matrix by computing the pairwise Euclidean distance:
%
% \begin{align}
%     & M \in \mathbb{R}^m \times \mathbb{R}^m \mid M_{ij} = ||v_i - v_j||_2\\
%     & \lim_{k\to\infty} M = D
% \end{align}
%
% where $D$ represents the true Euclidean distance matrix.
%
% while also collecting the necessary statistical information to standardize the target values in the output space (see "Algorithm" for more details).
It is worth noting that measuring distances on the contextual information space, as done by \citet{shah2024leaving}, introduces a bias in the surrogates, since in a stochastic setting the same $x$ input might be associated with different realizations and consequently with different loss landscapes.

We use the discussed similarity score to determine how relevant the data from one sample is for another sample.
% As a final step, to minimize the noise introduced by external samples, we rescale point-wise correlations for data coming from external surrogates.
This is achieved by operating on the covariance matrix of each GP, by downscaling the kernel outputs.
Specifically, let $\hat{y}'_i$ and $\hat{y}''_j$ be two predictions collected respectively for the training examples $i$ and $j$; then we have:
\begin{equation}
    K_{new}(\hat{y}_i', \hat{y}_i'') = \frac{K_{old}(\hat{y}_i', \hat{y}_j'')}{1 + e^{\alpha} \|v_{i} - v_{j}\|_2}
\end{equation}
where $\alpha$ is a learnable scale factor and the kernel value is unaltered if $i = j$, since the corresponding distance is 0.
% , while $i$ and $j$ identify, among the $m$ training instances, the ones from which $\hat{y}'$ and $ \hat{y}''$ have been sampled from. By design, $M$ is a hollow symmetric matrix, hence $M_{ii}=0$, so $K$ is not altered where points belong to the same instance.
%
The collection phase required for this sample sharing process is also useful to initialize the set of GPs, and to perform standardization of the regret values that are used for their training.

\changed{It is worth noting that, in principle, using a single GP surrogate with access to the contextual input $x$, rather than one surrogate per example, would enable data sharing in a more natural fashion.
Unfortunately, such a design choice would have a severe adverse effect on the surrogate scalability, as discussed in \cref{appendix:gp_scalability}.}

\paragraph{Algorithm}

We now dive into the concrete implementation of the main training procedure (see \cref{alg:training loop}), starting from the outer training loop to solve \cref{eq:sample-based-training}, and then moving to the inner training to optimize the GP surrogates.

Our surrogate loss should be combined with a fallback method, ideally one suitable for a general setting where no restrictive assumption on the regret function is made.
One such example is the Score Function Gradient Estimation (SFGE) approach by \citet{silvestri2023score}.
Using this method has two additional benefits.
First, SFGE also relies on stochastic smoothing, so that the set of predictions used for training the GPs can be naturally populated, their distribution of origin is known when computing $q(\hat{y})$, and the loss is semantically consistent between the surrogate and the fallback method.
% inherently applies stochastic smoothing to the loss function, so the $Y'$ and $Q$ sets are naturally populated by $(\hat{y}, r_i(\hat{y)})$ samples and their normal distributions of origin.
Second, the compound smoothing achieved by our GPs when few samples have been collected tends to compensate for the high variance and slow convergence of SFGE.
% As a second point, slow convergence is a strong limitation of SFGE, due to its high variance, but GP models reduce this variance in a twofold way: they operate on a more sample efficient smoothing, supported by importance sampling, and they apply another inner smoothing, coming from data interpolation. The first kind of smoothing decreases variance as the number of data points increases, while the second one is stronger when few points are available. This mechanism leads to informative gradients independently from how many training points have been collected by the GP. 
That said, it should be possible to use our surrogate loss with a different fallback method, such as those from \citet{hu2023two,elmachtoub2022smart} or \cite{huang2024decision}.
\changed{This investigation is left for future research.}

\begin{algorithm}[tb]
    \caption{\bf Training loop -- gradient computation}\label{alg:training loop}

    \begin{algorithmic}

        \STATE $gp \gets \texttt{initializeGPs}(y)$
        \FOR{\texttt{epoch in EPOCHS}}
            \STATE $loss \gets 0$
            \STATE $i \gets 0$
            \WHILE{$i < m$}
                \STATE $\hat{y}_i \gets h_\theta(x_i)$
                \STATE $ \bar{r}_i, \sigma_i \gets gp_i(\hat{y}_i)$
                \IF{$\sigma_i < \beta$}
                    \STATE $loss \gets loss + \bar{r}_i$
                \ELSE
                    \STATE $q_i \gets \mathcal{N}(\hat{y}_i, \sigma)$
                    \STATE $\hat{y}_i \sim q_i$
                    \STATE $gp_i\texttt{.add}(\hat{y}_i, r_i(\hat{y}_i), q_i)$
                    \STATE $loss \gets loss + \texttt{SFGE}(\hat{y}_i, r_i(\hat{y}_i), q_i)$
                \ENDIF
                \STATE $i \gets i + 1$
            \ENDWHILE
            \STATE $gradient \gets loss\texttt{.backward()}$
        \ENDFOR
      
    \end{algorithmic}
\end{algorithm}

The first step in \cref{alg:training loop} is a pretraining stage where we initialize the GPs and we collect pairs $(\hat{y}, r_i(\hat{y}))$ via LHS.
We scale the number of points to be sampled logarithmically with respect to the dimensionality of $D_y$. We use these points to compute statistics to normalize the input $\hat{y}$ (as GPs expect 0-mean input values) and to standardize the output $r_i(\hat{y})$.
Then, for each training example, we employ the associated GP to compute a predicted mean and standard deviation. If the latter is below a threshold $\beta$ (i.e., the GP is confident with respect to $\beta$), we take the mean as a surrogate loss and differentiate through the GP; otherwise, we call the SFGE procedure and we add the generated sample (and its generating distribution) to the corresponding GP. We aggregate loss terms for each training instance into a single loss, mixing values from the surrogates and the fallback method, before gradient computation.
% ;
% in fact, the resulting gradients are consistent, since both approaches rely on the same form of stochastic smoothing.
% We show the pseudo-code for gradient computation in \cref{alg:training loop}.

% Even if the two methods produce non-comparable values, we can proceed without an explicit conversion, as SFGE is already a gradient approximator of the smoothed function. 

We trigger surrogate training when at least $t$ new sample-target pairs are available. New data points are then shared between similar GPs, if such option is enabled, according to a maximum tolerable Euclidean distance $d_{max}$.
We replace the $r_i(\hat{y})$ targets with the smoothed ones $\bar{r}_i(\hat{y})$, using \cref{eq:importance sampling} on the collected samples and distributions.
We also apply data preprocessing by normalizing the inputs and standardizing the outputs, on the basis of the statistical information extracted in the pretraining phase.
We train GPs by maximum likelihood estimation, with a classical RBF kernel and a length-scale prior $l_i \sim \mathit{LogNormal}(\nicefrac{\log(\mathrm{dim}(Y))}{2}, 1)$. We add this last regularization term, following \citet{hvarfner2024vanilla}, to improve scalability in higher dimensions.
Finally, the RBF kernels are warm-started in all subsequent trainings, so as to speed up their training.
The code for our method is publicly available at \url{currently-in-the-supplemental-material}.

% \paragraph{Main Ideas}

% \begin{itemize}
%     \item Main idea: speed by using a trainable surrogate
% \end{itemize}

% \paragraph{Surrogate Model}

% Surrogate:
% \begin{equation}
%     \tilde{r}_i : D_y \rightarrow \mathbb{R}
% \end{equation}
% For a given realization $y$, the surrogate maps a prediction vector $\hat{y}$ into a corresponding loss term.

% \begin{itemize}
%     \item Differentiable
%     \item Unbiased
%     \item Provide info on its reliability
% \end{itemize}

% \paragraph{Stochastic Smoothing}

% Smoothed $r$:
% \begin{equation}
%     \bar{r}: D_y \rightarrow \mathbb{R}
% \end{equation}
% and:
% \begin{equation}
%     \bar{r}(\hat{y}) = \mathrm{E}_{\hat{y}' \sim \mathcal{N}(\hat{y}, \sigma)} \left[
%         r(y, \hat{y}')
%     \right]
% \end{equation}

% Rather than actually sampling from a Normal distribution centered on a given $\hat{y}$, we approximate the expectation via Importance Sampling.
% Given a sample...

% \begin{itemize}
%     \item Unbiased $\Rightarrow$ 0-gradient
%     \item Stochastic smoothing
% \end{itemize}

% \paragraph{Data Sharing}

% \begin{itemize}
%     \item Mechanism
%     \item Analogy with the second order terms in Taylor expansion
% \end{itemize}

% \paragraph{Fallback Method and Algorithm}

% \begin{itemize}
%     \item Fallback method (SFGE)
%     \item Pseudo-code
% \end{itemize}

% \paragraph{Surrogate Model Training}

\section{Experiments}

We now discuss the experimental analysis conducted to assess the robustness, reliability, accuracy, and especially scalability of our method. We designed experiments to answer four main research questions. \textbf{Q1.} How does our surrogate loss perform in terms of decision quality compared to the relevant baselines? \textbf{Q2.} How many calls to the black-box solver does it require? \textbf{Q3.} How much runtime does it take to converge compared to the other methods? \textbf{Q4.} Can it be scaled to high dimensions?
% To analyze the quality of our approach we tested it in diverse benchmarks against state-of-the-art methods in DFL.

\paragraph{Benchmarks}

We consider three problem classes, many of which include recourse actions to repair violated constraints, thus leading to misspecified, realistic, decision problems.

% We employ three classes of problems in our experiments:

\textit{Knapsack (KP).} We generate 1-0 KP datasets following the procedure proposed by \citet{elmachtoub2022smart} to model a stochastic mapping between input features $x$ and ground-truth targets $y$, with a polynomial degree $deg = 5$, number of input features $dim(X) = 5$, a noise half-width $\bar{\epsilon}= 0.5$, and a Poisson distribution to correlate $x$ and $y$ . In our experiments we build datasets on a target space of $dim(Y) = 50$ items. We make use of three different setups, respectively injecting uncertainty (i.e., stochastic correlation) into weights, values and capacity. We also adopt the recourse action system by \citet{silvestri2023score}, with a fixed penalty $p=10$.

\textit{Weighted Set Multi-Cover (WSMC).} We create WSMC datasets with $dim(X)=5$ input features, $dim(Y) = 10$ items and $s=50$ sets, following the guidelines by \citet{grossman1997computational} to generate realistic availability matrices and the same stochastic correlation as in \citet{elmachtoub2022smart}. We use the recourse action adopted by \citet{silvestri2023score}, with a fixed penalty $p=10$.

\textit{Toy.} We define a synthetic toy dataset. In this setting the input features are deterministically mapped to a target space, using a weight matrix $W \in R^{dim(Y) \times dim(X)} \sim U(0, 1)$, such that $y = Wx$. The underlining optimization problem is a trivial map where $z^*(\hat{y}) = \hat{y}$, while the cost is given by a pseudoconvex piecewise step function with minimum in $y$: $g(y, z^*(\hat{y})) = s \lfloor \nicefrac{\|y - \hat{y}\|_2}{l}\rfloor$, where $s$ controls the step heights and $l$ determines the distance between steps. We set $s = 5$ and $l=1$ in our experiments.
This new benchmark provides a controlled way to investigate convergence properties, as the entire loss landscape is known. Additionally, it allows for changing the dimensionality of $Y$, while keeping a negligible cost for computing $z^*$, so to stress DFL techniques on larger scales.

\paragraph{Baselines}

Our baselines include a predictive model trained for maximum likelihood, referred to as Prediction Focused Learning (PFL), plus state-of-the-art DFL methods that apply to black-box settings.
% as a
% As a guiding baseline, useful to set reference loss values, we use the classical PFL.
In particular, we include in our comparison the SFGE method by \citet{silvestri2023score}, which also serves as fallback for our approach, allowing us to directly assess the impact of the surrogate-based acceleration.
% reference for accuracy and runtime when no acceleration is provided.
For this method, we use a trainable parameter for $\sigma$, with starting value 0.1, and we warm start the predictor via PFL training.
% $\sigma=0.1$ as a trainable parameter for SFGE and for our method; in addition backbones are pretrained via PFL in both cases.
We then consider EGL, LODL, and LANCER, respectively from \citet{shah2024leaving}, \citet{shah2022decision} and \citet{zharmagambetov2023landscape}, as representative of other state-of-the-art black-box surrogate-based approaches.
We employ the same set of hyperparameters proposed by the authors, implementing all the four convex surrogates (MSE, Directed-MSE, Quadratic, Directed-Quadratic) and fixing the number of samples to $250$ for LODL and to $32$ or $42$ for EGL, to put it on par with our model in terms of calls, for a fair regret comparison. We adopt the same surrogate model architecture proposed in \citet{zharmagambetov2023landscape} for LANCER, with two hidden layers of $200$ units and tanh activation functions; we set $t=10$ for the dual training.
As a representative of a state-of-the-art DFL method requiring restrictive assumptions, we include SPO+ by \citet{elmachtoub2022smart}, where applicable.

\paragraph{Results}

\begin{figure}[tbh]
    \centering
    \includegraphics[width=0.9\textwidth]{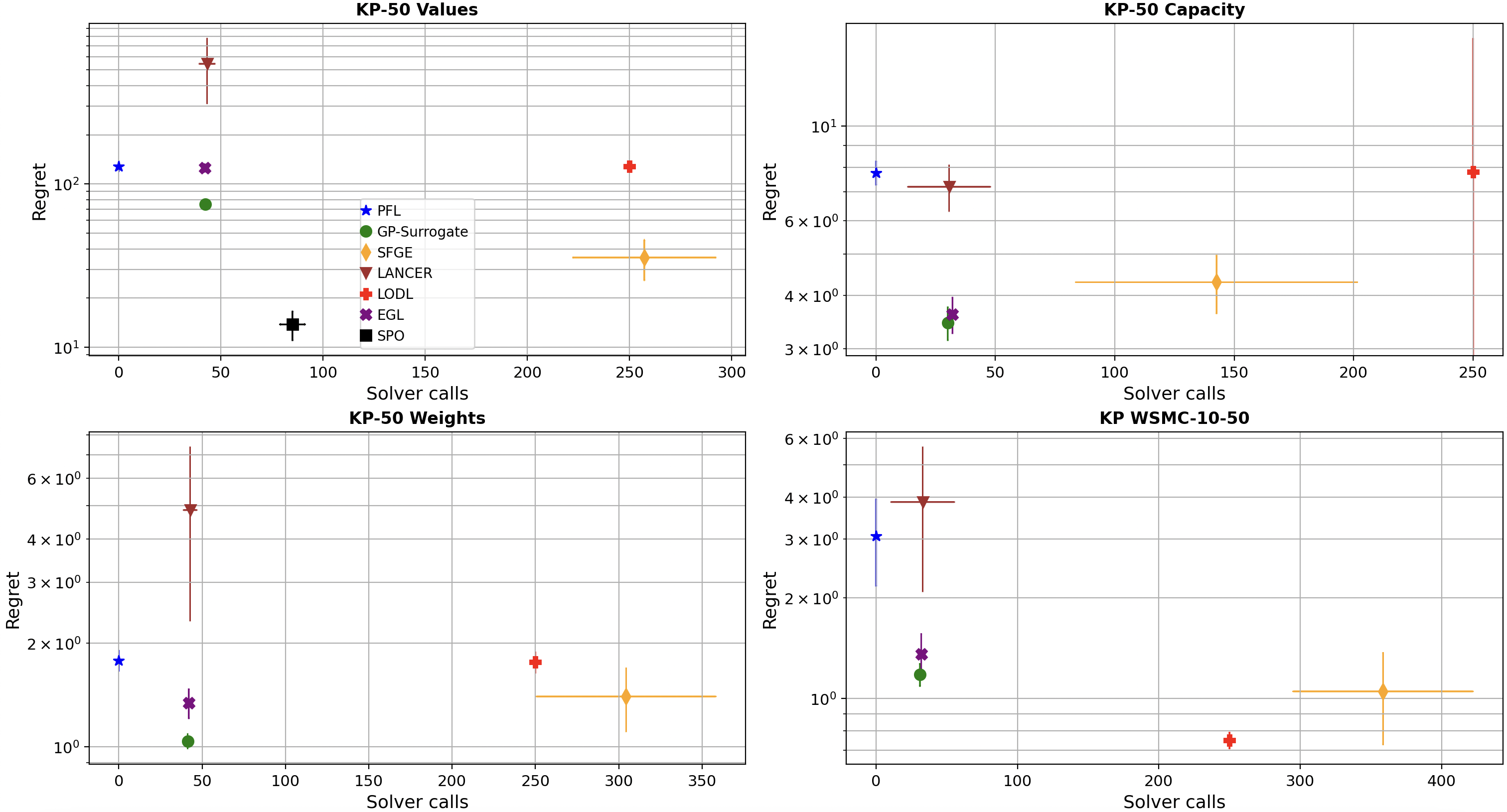}
    \caption{Regret and solver calls across the benchmarks: points represent the average regret (on a logarithmic scale) and average solver calls per instance for all the baselines. Lines represent standard deviations for each method. Only the best (lowest regret) models are reported for LODL and EGL.}
    \label{fig:regret_calls_plot}
\end{figure}

In this section we report results on a set of experiments designed to answer the discussed research questions. For each setting we generate $5$ datasets, each one containing 1000 instances, differently split into train ($80\%$), validation ($10\%$) and test sets ($10\%$).
All the models are trained with the Adam algorithm by \citet{kingma2014adam} and a learning rate $lr=10^{-3}$, using an early stopping criterion on regrets computed on the validation set, to avoid overfitting.
We enable stochastic smoothing, differentiation, and pretraining for our method, here named \ouracronym{}.
By default, we keep sample sharing off, as it is best suited to improve speed on very time-consuming problems, at the cost of solution quality
% that might worsen accuracy, so it is suited for extremely time-consuming problems
(see \cref{table:ablation} for more details).
We set $\beta=1.0$ as a tradeoff between speed and precision in learning, as shown in \cref{appendix:sensitivity analysis}, and $t=40$.
All the experiments were run on an Apple M3 Pro CPU with 12 cores and 18GB of RAM.

\textbf{Q1}. We compare the regret for each approach on the KP with $50$ items (\textsc{KP-50}) and uncertainty in weights, values and capacity, and on the the WSMC with $10$ items and $50$ sets (\textsc{WSMC-10-50}).
Results, summarized in \cref{fig:regret_calls_plot} and presented in extended form in \cref{appendix:extended results}, indicate that using our surrogate on top of SFGE leads to solutions of similar quality (or even better) in all but one benchmark, and with less variability.
% loss has a competitive accuracy compared to its fallback method (SFGE), even outperforming it in two settings.
% This is the most relevant comparison, as \ouracronym{} depends on it, and the goal is to accelerate it.
\ouracronym{} provides better regret and stability than the best approaches in the LODL and EGL class.
We also found the performance of LODL and EGL to be inconsistent w.r.t. the employed convex surrogate, as highlighted in \cref{appendix:extended results}.
% Furthermore, these models require to implement and train multiple surrogate losses before identifying the best one.
Overall, our method seems to provide a better alternative for accelerating training than these approaches, at least in terms of regret, most likely thanks to its lack of a structural estimation bias, and to the ability to switch to the fallback method in case of low confidence.
% We highlight how \ouracronym{} acts similarly to these models, but its reliability and consistency are likely higher as it is an unbiased estimator and it provides a confidence score.
The surrogate model used by LANCER seems often unable to reasonably approximate the real loss landscape, causing inconsistent training results and high regrets (even more than PFL), despite a significant reduction in terms of solver calls.
The SPO method, when applicable, significantly outperforms all the others, which suggests that traditional DFL approaches should still be preferred to black-box ones when permitted by their assumptions.
% As a final note, SPO clearly outperforms all the other methods, but it is limited by its domain assumptions.

\textbf{Q2.} We adopt the same datasets as in Q1 to analyze the computational cost of all approaches.
Results are again depicted in \cref{fig:regret_calls_plot}. In \cref{appendix: calls comparison} we report the number of solver calls (per training sample), and the total runtime for each method.
The former metric should be considered more important, as for any sufficiently difficult problem the decision time will be the dominant factor.
% We consider first metric more informative, since typically a solver call is the most time-demanding operation, setting a $O(n)$ time complexity.
\ouracronym{} reduces the number of calls by almost an order of magnitude compared to its fallback method, greatly increasing viability in a practical setting.
% Also in this case, results prove that \ouracronym{} effectively reduce the calls to the solver, accelerating its fallback method by almost one order of magnitude.
We align the number of EGL samples to the solver calls from our method and observe EGL performing significantly worse in terms of regret.
The LODL approaches, in their reference configuration, perform more solver calls for even worse regret in all but one benchmark.
% In addition, calls are close to those by LODL and EGL models, despite the much lower average regret examined above.
The same considerations apply for the runtime.
% We finally point out that our surrogate parameters can be saved to warm start subsequent rounds of training.
% can be saved and reloaded after each training (and pretraining), saving most of the computational time whenever multiple trainings are needed.

\textbf{Q3.} We evaluate scalability for complex problems with time-consuming solver calls, by building versions of the WSMC-10 datasets with an increasing number of item sets.
% (i.e. runtime scales linearly with the available sets).
For this experiment we set a time limit of $900s$ per training attempt, to simulate real-world scenarios with limited computational resources.
We compare \ouracronym{} with SFGE and report the results in \cref{table:q3 runtime}.
As it can be seen, \ouracronym{} consistently achieves training convergence.
SFGE fails to do so even in the simplest cases, which causes the approach to have worse regret than \ouracronym{} in this case.
% \Cref{table:q3 runtime} shows the practical benefits of \ouracronym{}, which converges within the fixed time limit in all of the settings; while SFGE does not converge in time even for the simplest case.

\begin{table*}[bt]
\centering
\caption{Runtime and average regret for \ouracronym{} and SFGE on WSMC-10 with 250, 500, 750 and 1000 sets. Time limit is $900s$.}
\label{table:q3 runtime}

\small  
    \begin{tabular}{clcccc}
    
    \toprule
    
    & \textit{Method} & \textit{WSMC-10-250} & \textit{WSMC-10-500} & \textit{WSMC-10-750} & \textit{WSMC-10-1000} \\

    \midrule

    %\multicolumn{5}{c}{Runtime} \\
    
    %\midrule

        \multirow{2}{*}{\textit{Runtime}}
        %\textsc{PFL}  & $ m \pm s $ & $ m \pm s $ & $ m \pm s $ & $ m \pm s $ \\
        & \ouracronym{}  & $ \mathbf{289.24 \pm 6.54} $ & $ \mathbf{459.70 \pm 15.73} $ & $ \mathbf{638.50 \pm 14.70} $ & $ \mathbf{818.44 \pm 17.74} $ \\
        & \textsc{SFGE}  & $ 900.0^* \pm 0.0 $ & $ 900.0^* \pm 0.0 $ & $ 900.0^* \pm 0.0 $ & $ 900.0^* \pm 0.0 $ \\

    \midrule

    %\multicolumn{5}{c}{Average regret} \\
    
    %\midrule

        \multirow{2}{*}{\textit{Regret}}
        %\textsc{PFL}  & $ m \pm s $ & $ m \pm s $ & $ m \pm s $ & $ m \pm s $ \\
        & \ouracronym{}  & $ \mathbf{1.55 \pm 0.57} $ & $ \mathbf{1.51 \pm 0.50} $ & $ \mathbf{1.48 \pm 0.54} $ & $ \mathbf{1.51 \pm 0.59} $ \\
        & \textsc{SFGE}  & $ 1.83 \pm 0.86 $ & $ 2.44 \pm 1.05 $ & $ 2.82 \pm 1.20 $ & $ 3.54 \pm 1.25 $ \\
        
    \bottomrule
    \end{tabular}

\end{table*} 

\textbf{Q4.} We evaluate scalability for high-dimensional predictions (as opposed to larger-size problems like in Q3), by increasing the number of decision variables in the Toy benchmark.
% using the Toy benchmark with increasingly higher dimensions.
We report results in \cref{table:q4 dimensions}. SFGE gets stuck into local optima, likely because of the nature of this synthetic problem, which requires strong variations of $\sigma$ across the training steps.
Conversely, \ouracronym{} is still able to effectively minimize the loss function.
This behavior shows that our surrogates can enable convergence to high-quality solutions, even in high-dimensional spaces.
We conjecture this is partly due to the fact that, since the realization $y$ is always used when training the GPs, they can naturally identify the presence of a local minimum for $r_i$ at that location;
this property is shared with the surrogates from \citet{shah2022decision}, but with all the discussed benefits of our solution.
% its natural smoothing is sufficient even in high dimensions to converge: few points, including the ground-truth $y$, still allow to approximate a function with minimum in $y$, essentially reducing to the same surrogate logic of \citet{shah2022decision}, but with all the discussed benefits from our method.
\begin{table*}[tb]
\centering
\caption{Average regret for \ouracronym{} and SFGE on the Toy dataset with 64, 128, 256 and 512 dimensions.}
\label{table:q4 dimensions}

\small  
    \begin{tabular}{clcccc}
    
    \toprule
    
    & \textit{Method} & \textit{Toy-64} & \textit{Toy-128} & \textit{Toy-256} & \textit{Toy-512} \\
    
    \midrule

        \multirow{2}{*}{\textit{Regret}}
        %\textsc{PFL}  & $ m \pm s $ & $ m \pm s $ & $ m \pm s $ & $ m \pm s $ \\
        & \ouracronym{}  & $ \mathbf{5.61 \pm 2.97} $ & $ \mathbf{4.18 \pm 0.96} $ & $ \mathbf{1.29 \pm 0.46} $ & $ \mathbf{0.29 \pm 0.23} $ \\
        & \textsc{SFGE}  & $ 39.78 \pm 3.06 $ & $ 58.61 \pm 1.02 $ & $ 113.28 \pm 3.41 $ & $ 271.82 \pm 9.49 $ \\
        
    \bottomrule
    \end{tabular}

\bigskip

\caption{Average regret and average calls per instance for \ouracronym{} with different components on KP-50 with uncertain weights, values and capacity and WSMC-10 with 50 sets.}
\label{table:ablation}

\small  
    \begin{tabular}{clcccc}
    
    \toprule
    
    & \textit{Method} & \textit{KP-50 weights} & \textit{KP-50 values} & \textit{KP-50 capacity} & \textit{WSMC-10-50} \\

    \midrule

    % \multicolumn{5}{c}{Average regret} \\
    
    % \midrule

        \multirow{5}{*}{\rotatebox{90}{\textit{Regret}}}
        & \textsc{Full model}  & $ 1.14 \pm 0.11 $ & $  106.38 \pm 10.19 $ & $  3.52 \pm 0.33 $ & $  1.39 \pm 0.23 $ \\
        & \textsc{Smoothing OFF}  & $ 1.40 \pm 0.11 $ & $  149.67 \pm 16.68 $ & $  3.64 \pm 0.52 $ & $  1.42 \pm 0.19 $ \\
        & \textsc{Pretrain OFF}  & $ 1.15 \pm 0.08 $ & $  \mathbf{57.83 \pm 2.61} $ & $  5.04 \pm 0.33 $ & $  \mathbf{0.67 \pm 0.10} $ \\
        & \textsc{Sample sharing OFF}  & $ \mathbf{1.04 \pm 0.05} $ & $  75.07 \pm 5.20 $ & $  \mathbf{3.45 \pm 0.31} $ & $  1.18 \pm 0.09 $ \\
        & \textsc{Differentiation OFF}  & $ 1.63 \pm 0.21 $ & $  156.56 \pm 25.23 $ & $  3.61 \pm 0.57 $ & $  1.58 \pm 0.71 $ \\

    \midrule

    %\multicolumn{5}{c}{Average calls} \\

    %\midrule

        \multirow{5}{*}{\rotatebox{90}{\textit{Slv. calls}}}
        & \textsc{Full model}  & $ \mathbf{39.84 \pm 2.51} $ & $  \mathbf{35.41 \pm 0.48} $ & $  \mathbf{6.93 \pm 0.14} $ & $  \mathbf{28.31 \pm 1.10} $ \\
        & \textsc{Smoothing OFF}  & $ 41.58 \pm 0.74 $ & $  36.44 \pm 0.35 $ & $  6.97 \pm 0.12 $ & $  29.20 \pm 0.59 $ \\
        & \textsc{Pretrain OFF}  & $ 121.71 \pm 3.71 $ & $  88.34 \pm 2.55 $ & $  48.91 \pm 0.60 $ & $  72.65 \pm 9.41 $ \\
        & \textsc{Sample sharing OFF}  & $ 41.33 \pm 1.22 $ & $  42.32 \pm 0.09 $ & $  30.04 \pm 0.05 $ & $  31.06 \pm 0.35 $ \\
        & \textsc{Differentiation OFF}  & $ 42.40 \pm 4.69 $ & $  36.28 \pm 0.48 $ & $  6.95 \pm 0.15 $ & $  28.95 \pm 1.03 $ \\
        
    \bottomrule
    \end{tabular}

\end{table*} 

Overall, the experiments indicate that our approach can outperform its fallback method in terms of solver calls and total runtime, thus making it scalable to complex problems, while keeping a comparable and sometimes better decision quality.
The degree of acceleration is similar to the LODL, EGL, and LANCER models, but with improved consistency, reliability, and solution quality.
Finally, the method scales much better then SFGE on high-dimensional parameter spaces, thanks to the combination of stochastic and GP smoothing.

\paragraph{Ablation studies}
To prove the effectiveness of all the major components of our method, we also conducted an ablation study by separately disabling stochastic smoothing, pretraining, sample sharing and GP differentiation. Results, reported in \cref{table:ablation}, reveal that in all the settings removing differentiation or smoothing causes higher average regrets and solver calls, highlighting their relevance. Results with no sample sharing are the same of \cref{fig:regret_calls_plot}; we note that sharing points between GP models affects negatively the average regrets, but reduces the average calls. In most cases, this extra source of acceleration is not enough to justify the lower decision quality, but it can be extremely valuable in some cases, as observed in the KP-50 capacity benchmark where the number of solver call is almost two orders of magnitude lower than SFGE. For what concerns pretraining, the number of solver calls grows sensibly when it is turned off, proving its fundamental role for learning confident surrogates in early stages. However, relying more on the fallback method does not imply a downgrade in terms of regrets; in fact, in two settings out of four, we see a strong improvement when the surrogate usage is more moderate.

% \paragraph{Sensitivity analysis}
% In \cref{table:sensitivity} we analyze how $\beta$ influences results. As expected, higher values lead to an increase in the surrogate loss exploitation. However, we observe a counter-intuitive behavior in average regrets, which improve even if less confident estimations take the place of real regrets. We believe these results my be determined by spurious local optima introduced by the GP loss approximation.

% \input{sections/tables/sensitivity}

\section{Conclusions}

We present an approach to improve the applicability of DFL methods, targeting scenarios where solution and cost computation are time-consuming and where access to the problem structure and solver state is impossible or inconvenient.
In this setting, the existing DFL approaches are too slow to converge, they may not be applicable, or they accelerate the training process at the cost of a worse decision quality.
% Conversely, our method is applicable to general black-box optimization problems, and can lead to significant runtime reductions with little or no performance reduction.
We employ a GP-based surrogate loss function, trained online in alternation with a fallback method, to exploit the surrogate speed without loosing the ability to adapt to the regret function landscape.
% explore the predictive space whenever the surrogate confidence is low, and exploit its fast loss estimation otherwise.
We solve the 0-gradient issue relying on stochastic smoothing via importance sampling -- which also motivates our choice of SFGE for the fallback method.
% for this reason we combine it with SFGE, which is based on the same smoothing principle.
Our experimental evaluation reveals that our surrogate matches or outperforms SFGE, while reducing the number of solver calls by up to two orders of magnitude, depending on the problem and the model configuration.
Our method improves over related approaches in terms of sample efficiency, solution quality, or both.
% We also observed higher solution quality and consistency compared to the other surrogate-based methods.
Finally, we show that the approach can be scaled to higher dimensions.
Some potential directions of improvement remain unexplored.
A point of particular interest is the possibility to adjust the smoothing factor, or the points where smoothed regret is computed, at training time -- without the need to collect additional samples.
Moreover, by adapting acquisition functions from classical Bayesian optimization, it might be possible to remove the need for a fallback method.
Finally, similar to the LODL, EGL, and LANCER approaches, our learned surrogates could be exported for the construction of new training sets or generally for approximating regret evaluation.

\bibliography{iclr2026_conference}
\bibliographystyle{iclr2026_conference}

\clearpage
\appendix
%\section{Appendix}
%You may include other additional sections here.

\section{Conditions for the Effectiveness of the Classical DFL Setting}
\label{appendix:dfl_classical}

Most classical Decision Focused Learning approaches focus on the following setting:
\begin{equation}
    \argmin_\theta \mathrm{E}_{x, y \sim P(X, Y)}\left[ r(y, \hat{y}) \right]
\end{equation}
where we have:
\begin{align}
    r(y, \hat{y}) & = y^T z^*(\hat{y}) - y^T z^*(y) \label{eq:classi_regret} \\
    z^*(\hat{y}) & = \argmin_z \left\{ \hat{y}^T z \mid z \in C \right\} \label{eq:classi_opt}
\end{align}
This ``classical'' setting differs from the one described in this paper in three key respects:
\begin{enumerate}
    \item The considered optimization problem has a linear cost function and a constraint set that does not depend on the uncertain parameters
    \item The uncertain parameters correspond to the coefficients of the linear cost function
    \item The true cost function ($g$ in our setup) is the same as that used in the optimization problem, i.e. \cref{eq:classi_regret} and \cref{eq:classi_opt} use the same cost function
\end{enumerate}
The assumptions in the classical setup are used by state of the art DFL methods to obtain good convergence speed and excellent solution quality.
However, the same assumptions also introduce some notable limitations.

We start by observing that, since the $y^T z^*(y)$ term in the regret expression does not depend on the predictions, we have that:
\begin{equation}
    \argmin_\theta \mathrm{E}_{x,y \sim P(X, Y)} \left[ r(y, \hat{y}) \right] =
    \argmin_\theta \mathrm{E}_{x,y \sim P(X, Y)} \left[ y^T z^*(\hat{y}) \right]
    \label{eq:things_go_south}
\end{equation}
A similar equivalence holds for all DFL settings, including the one we consider.
In the classical DFL setting, however, the linearity of the cost function also implies that:
\begin{align}
    \argmin_\theta \mathrm{E}_{x,y \sim P(X, Y)} \left[ y^T z^*(\hat{y}) \right] =
    \argmin_\theta \mathrm{E}_{x \sim P(X)} \left[ \mathrm{E}_{y \sim P(Y \mid x)} \left[ y \right]^T z^*(\hat{y}) \right]
    \label{eq:the_party_is_over}
\end{align}
It can be seen that an optimal solution for \cref{eq:the_party_is_over} can be obtained by setting $\hat{y} = \mathrm{E}_{x,y \sim P(X, Y)} \left[ y \right]$, i.e. if the ML predictor provides a correct estimate of the expectation of the uncertain parameters.

In practical terms, this means that, in the classical setting, \emph{the advantage of DFL over PFL becomes vanishingly small, unless the ML predictor has an irreducible estimation error}, i.e. it is misaligned -- see for example \cite{huang2024decision}.
DFL methods can still be useful, for example when the correct distribution $P(Y \mid X)$ is not known, or when a simpler ML model (e.g. a linear regressor) is desirable for easier analysis.
However, this weakness significantly reduces the practical appeal of the classical DFL setting.

The reduction from \cref{eq:things_go_south} does not apply in general to our black-box setting,  which makes our approach capable of addressing stochastic problems with non-linear costs and/or recourse action, on which the advantage provided by DFL can be robust, regardless of the predictor expressivity.

\section{Estimation Bias Analysis} 
\label{appendix:estimation_bias}

One of the appeals of our proposed approach is that it relies on surrogates that can provide an unbiased estimation of the regret.
While this claim is limited to the abundant data regime, it is still relevant since it implies that our approach does not introduce an irreducible error in the regret estimation.
Here, we proceed to detail our claim and provide a proof sketch.

Consider the DFL training problem as stated in \cref{eq:sample-based-training}, and its variant defined over our proposed surrogates, i.e.:
\begin{equation}
    \argmin_\theta \frac{1}{m} \sum_{i=1}^m \tilde{r}_i(\hat{y})
\end{equation}
The loss functions employed in the two cases are separable over the training examples.
As a consequence, one can see that as long as the surrogate $\tilde{r}_i(\hat{y}_i)$ can provide an arbitrarily accurate approximation of the true regret $r(y_i, \hat{y}_i)$ for every input point, then the surrogate loss is also an arbitrary accurate approximation of the true loss.
Formally, we have that:
\begin{equation}
    \forall i=1..m, \hat{y}_i \in D_y, \tilde{r}_i(\hat{y}_i) \simeq r(y_i, \hat{y}_i)
    \implies
    \forall \hat{y}_i \in D_y,
    \sum_{i=1}^m \tilde{r}_i(\hat{y}_i) \simeq \sum_{i=1}^m r(y_i, \hat{y}_i)
\end{equation}
By relying on linearity of differentiation operator and on universal quantification over the regret input, we get an analogous result for the loss gradient:
\begin{equation}
    \forall i=1..m, \hat{y}_i \in D_y,
    \tilde{r}_i(\hat{y}_i) \simeq r(y_i, \hat{y}_i)
    \implies
    \forall \hat{y}_i \in D_y,
    \nabla_\theta \sum_{i=1}^m \tilde{r}_i(\hat{y}_i) \simeq \nabla_\theta \sum_{i=1}^m r(y_i, \hat{y}_i)
\end{equation}
which can be proved by relying on the definition of partial derivative as the limit for a finite difference.
Therefore, as long as each surrogate $\tilde{r}_i(\hat{y}_i)$ can approximate sufficiently well the regret landscape for the corresponding training example, then our surrogate loss accurately approximates the true DFL loss and its gradient accurately approximates the true gradient.

Now, it can be seen the true regret $r(y_i, \hat{y}_i)$ is a deterministic function of $\hat{y}_i$, since it refers to a specific realization of the random variable $Y$.
Assuming arbitrarily abundant, uniformly sampled, values of $\hat{y}_i$, a Gaussian Process with an RBF kernel can approximate any deterministic function with an approximation error that converges to 0 as the number of datapoints used for training the GP grows.
If the regret landscape $r(y_i, \hat{y}_i)$ is smooth, then the approximation error can reach 0;
for the kind of discontinuous, piecewise constant, regret landscapes that we are primarily interested in, the approximation error approaches $0^+$ rather than 0, which is sufficient for our purpose.

In our approach, the GP surrogates are trained to approximate the smoothed regret $\bar{r}_i(\hat{y})$, rather then the true regret.
We will show that, under some conditions, this can still result in an arbitrarily accurate estimate of the true regret, in the sense that the approximation error can approach $0^+$.
First, observe that, in the same abundant data regime considered before and for a centered smoothing distribution (such as a the Normal distribution we use), the expectation used in stochastic smoothing will be estimated to arbitrarily accurate precision by a sample average.
Second, for a sufficiently low value of $\sigma$ the Normal distribution will converge to a Dirac delta, and as a consequence:
\begin{equation}
    | \bar{r}_i(\hat{y}) - r(y_i, \hat{y}_i) | \xrightarrow[\sigma \to 0^+]{} 0^+
\end{equation}
Therefore, an arbitrarily accurate estimate of the true regret can be obtained by lowering the smoothing factor $\sigma$.

Overall, we have that: 1) for abundant data points sampled uniformly at random; and 2) for a sufficiently low smoothing factor $\sigma$, our surrogates can provide an arbitrarily accurate approximation of the true regret.

From a practical point of view, using abundant samples runs counter to our main goal of improving scalability, and using larger values of $\sigma$ will typically provide more informative gradient.
The result we have just proved is however significant since it means our surrogate \emph{does not suffer from an irreducible approximation error}, unlike other approaches from the literature such as LODL or EGL.

\section{Scalability Considerations}
\label{appendix:gp_scalability}

Due to the complexity of the DFL pipeline, there are different terms that can affect scalability.
Those include:
\begin{itemize}
    \item The number $M$ of training examples
    \item The number $H$ of training epochs
    \item The number $m_i$ of datapoints (prediction-regret pairs) used to train the GP surrogate for the $i$-th example
    \item The number $n_y$ of predicted parameters appearing in the optimization model (and representing the input of our GP surrogates)
    \item The number $n_x$ of components in the contextual information vector
\end{itemize}
Since our focus is on improving training time scalability, we can expect that, in a practical situation:
\begin{itemize}
    \item The number of parameters $n_y$ should be relatively small, since the optimization problem being solved will likely have NP-hard or high-degree polynomial complexity
    \item The number of datapoints $m_i$ for each example should be quite small, and in particular much smaller than the number of epochs $H$, otherwise using only the fallback method would be better
\end{itemize}
The number of examples $M$, of training epochs $H$, and of components $n_x$ in the contextual information vector can be large in general.

We can now observe that:
\begin{itemize}
    \item Since we learn a separate GP per example, the computational cost for our surrogates scales as $O(M)$ with respect to the number of examples
    \item The cost for computing the covariance matrices used by each GP scales as $O(n_y m_i^2)$, since it requires $O(m_i^2)$ kernel computations, each running in $O(n_y)$. As stated above, both $n_y$ and $m_i$ can be expected to be small in practice
    \item The cost for training (and re-training) the GP surrogates via gradient descent can be expected to scale as $O(H)$ with respect to the number of epochs
\end{itemize}
Our sample sharing mechanism, if enabled, complicates the picture, by virtually increasing the number $m_i$ of datapoints used to training every GP.
However, since we use a distance cutoff for including shared samples, we observed only a modest increase in our experiments.

Overall, our design ensures that all higher-degree polynomial complexity terms are computed over low-dimensional objects, thus keeping the method quite scalable.

Notice that using a single surrogate model using pairs $(x, \hat{y})$ as input -- a seemingly more natural choice -- would allow to directly train the model over all collected datapoints, but would also make the covariance computation scale as $O((n_x + n_y) (m_i M)^2)$, which would be prohibitively expensive.

\section{Extended results} 
\label{appendix:extended results}

We report in \cref{table:accuracy extended} complete results relative to the average regret score for all the models, including all the LODL and EGL variations.

\begin{table*}[h]
\centering
\caption{Average regret for all the models on KP-50 with uncertain weights, values and capacity and WSMC-10 with 50 sets.}
\label{table:accuracy extended}

\small  
    \begin{tabular}{lcccc}
    
    \toprule
    
    \textit{Method} & \textit{KP-50 weights} & \textit{KP-50 values} & \textit{KP-50 capacity} & \textit{WSMC-10-50} \\
    
    \midrule
        
        \textsc{PFL}  & $ 1.78 \pm 0.12 $ & $ 128.55 \pm 8.99 $ & $ 7.77 \pm 0.50 $ & $ 3.06 \pm 0.89 $ \\
        \ouracronym{}  & $ \mathbf{1.04 \pm 0.05} $ & $ 75.07 \pm 5.20 $ & $ \mathbf{3.45 \pm 0.31} $ & $ 1.18 \pm 0.09 $ \\
        \textsc{SFGE}  & $ 1.40 \pm 0.29 $ & $ 35.40 \pm 9.77 $ & $ 4.30 \pm 0.67 $ & $ 1.05 \pm 0.32 $ \\
        \textsc{LANCER}  & $ 4.85 \pm 2.53 $ & $ 545.14 \pm 233.63 $ & $ 7.20 \pm 0.89 $ & $ 3.87 \pm 1.78 $ \\
        \textsc{LODL-MSE}  & $ 1.76 \pm 0.12 $ & $ 127.78 \pm 9.36 $ & $ 7.80 \pm 8.24 $ & $ 3.04 \pm 0.88 $ \\
        \textsc{LODL-Quadratic}  & $ 2.63 \pm 0.34 $ & $ 813.19 \pm 108.98 $ & $ 9.34 \pm 0.35 $ & $ 5.18 \pm 1.14 $ \\
        \textsc{LODL-Directed-MSE}  & $ 5.85 \pm 0.41 $ & $ 315.08 \pm 55.10 $ & $ 17.37 \pm 3.97 $ & $ \mathbf{0.75 \pm 0.04} $ \\
        \textsc{LODL-Directed-Quadratic}  & $ 2.06 \pm 0.21 $ & $ 190.34 \pm 5.84 $ & $ 14.96 \pm 0.41 $ & $ 2.92 \pm 0.89 $ \\
        \textsc{EGL-MSE}  & $ 1.78 \pm 0.09 $ & $ 125.48 \pm 6.69 $ & $ 6.41 \pm 1.02 $ & $ 2.25 \pm 0.32 $ \\
        \textsc{EGL-Quadratic}  & $ 1.34 \pm 0.13 $ & $ 247.49 \pm 95.72 $ & $ 6.28 \pm 0.33 $ & $ 2.43 \pm 0.93 $ \\
        \textsc{EGL-Directed-MSE}  & $ 1.75 \pm 0.12 $ & $ 129.74 \pm 3.76 $ & $ 6.13 \pm 0.37 $ & $ 2.32 \pm 1.42 $ \\
        \textsc{EGL-Directed-Quadratic}  & $ 1.97 \pm 0.06 $ & $ 275.98 \pm 70.70 $ & $ 3.61 \pm 0.35 $ & $ 1.36 \pm 0.20 $ \\
        \textsc{SPO}  & $ - $ & $ \mathbf{13.78 \pm 2.82} $ & $ - $ & $ - $ \\
        
    \bottomrule
    \end{tabular}

\end{table*} 

%\clearpage

\section{Solver calls and runtime comparison}
\label{appendix: calls comparison}
In \cref{table:q2 calls} we show the average solver calls per instance and the runtime for each baseline model.

\begin{table*}[h]
\centering
\caption{Average calls per instance and runtime for all the models on KP-50 with uncertain weights, values and capacity and WSMC-10 with 50 sets.}
\label{table:q2 calls}

\small  
    \begin{tabular}{clcccc}
    
    \toprule
    
    & \textit{Method} & \textit{KP-50 weights} & \textit{KP-50 values} & \textit{KP-50 capacity} & \textit{WSMC-10-50} \\
    
    \midrule
        
        % \textsc{PFL}  & $ 0.0 \pm 0.0 $ & $ 0.0 \pm 0.0 $ & $ 0.0 \pm 0.0 $ & $ 0.0 \pm 0.0 $ \\
        % \ouracronym{}  & $ 41.33 \pm 1.22 $ & $ 42.32 \pm 0.09 $ & $ 55.04 \pm 0.05 $ & $ 45.06 \pm 0.35 $ \\
        % \textsc{SFGE}  & $ 304.40 \pm 53.80 $ & $ 257.20 \pm 34.97 $ & $ 142.60 \pm 58.95 $ & $ 358.60 \pm 63.62 $ \\
        % \textsc{LODL-MSE}  & $ 250.0 \pm 0.0 $ & $ 250.0 \pm 0.0 $ & $ 250.0 \pm 0.0 $ & $ 250.0 \pm 0.0 $ \\
        % \textsc{LODL-Quadratic}  & $ 250.0 \pm 0.0 $ & $ 250.0 \pm 0.0 $ & $ 250.0 \pm 0.0 $ & $ 250.0 \pm 0.0 $ \\
        % \textsc{LODL-Directed-MSE}  & $ 250.0 \pm 0.0 $ & $ 250.0 \pm 0.0 $ & $ 250.0 \pm 0.0 $ & $ 250.0 \pm 0.0 $ \\
        % \textsc{LODL-Directed-Quadratic}  & $ 250.0 \pm 0.0 $ & $ 250.0 \pm 0.0 $ & $ 250.0 \pm 0.0 $ & $ 250.0 \pm 0.0 $ \\
        % \textsc{EGL-MSE}  & $ 32.0 \pm 0.0 $ & $ 32.0 \pm 0.0 $ & $ 32.0 \pm 0.0 $ & $ 32.0 \pm 0.0 $ \\
        % \textsc{EGL-Quadratic}  & $ 32.0 \pm 0.0 $ & $ 32.0 \pm 0.0 $ & $ 32.0 \pm 0.0 $ & $ 32.0 \pm 0.0 $ \\
        % \textsc{EGL-Directed-MSE}  & $ 32.0 \pm 0.0 $ & $ 32.0 \pm 0.0 $ & $ 32.0 \pm 0.0 $ & $ 32.0 \pm 0.0 $ \\
        % \textsc{EGL-Directed-Quadratic}  & $ 32.0 \pm 0.0 $ & $ 32.0 \pm 0.0 $ & $ 32.0 \pm 0.0 $ & $ 32.0 \pm 0.0 $ \\
        % \textsc{SPO}  & $ - $ & $ 85.0 \pm 6.09 $ & $ - $ & $ - $ \\

        % \midrule

        %\multicolumn{5}{c}{Average calls} \\

        %\midrule

        \multirow{5}{*}{\rotatebox{90}{\textit{Slv. calls}}}
        %\textsc{PFL}  & $ 0.0 \pm 0.0 $ & $ 0.0 \pm 0.0 $ & $ 0.0 \pm 0.0 $ & $ 0.0 \pm 0.0 $ \\
        & \ouracronym{}  & $ \mathbf{41.33 \pm 1.22} $ & $ 42.32 \pm 0.09 $ & $ \mathbf{30.04 \pm 0.05} $ & $ \mathbf{31.06 \pm 0.35} $ \\
        & \textsc{SFGE}  & $ 304.40 \pm 53.80 $ & $ 257.20 \pm 34.97 $ & $ 142.60 \pm 58.95 $ & $ 358.60 \pm 63.62 $ \\
        & \textsc{LANCER}  & $ 42.8 \pm 3.96 $ & $ 43.2 \pm 3.86 $ & $ 30.60 \pm 17.24 $ & $ 33.0 \pm 22.35 $ \\
        & \textsc{LODL (All)}  & $ 250.0 \pm 0.0 $ & $ 250.0 \pm 0.0 $ & $ 250.0 \pm 0.0 $ & $ 250.0 \pm 0.0 $ \\
        & \textsc{EGL (All)}  & $ 42.0 \pm 0.0 $ & $ \mathbf{42.0 \pm 0.0} $ & $ 32.0 \pm 0.0 $ & $ 32.0 \pm 0.0 $ \\
        & \textsc{SPO}  & $ - $ & $ 85.0 \pm 6.09 $ & $ - $ & $ - $ \\

        \midrule

        %\multicolumn{5}{c}{Runtime} \\

        %\midrule

        \multirow{5}{*}{\rotatebox{90}{\textit{Runtime}}}
        %\textsc{PFL}  & $ 0.0 \pm 0.0 $ & $ 0.0 \pm 0.0 $ & $ 0.0 \pm 0.0 $ & $ 0.0 \pm 0.0 $ \\
        & \ouracronym{}  & $ 199.40 \pm 6.77 $ & $ 136.01 \pm 10.02 $ & $ 274.52 \pm 66.50 $ & $ 228.73 \pm 20.59 $ \\
        & \textsc{SFGE}  & $ 717.28 \pm 127.26 $ & $ 774.16 \pm 173.62 $ & $ 396.93 \pm 160.09 $ & $ 1849.74 \pm 344.25 $ \\
        & \textsc{LANCER}  & $ 207.31 \pm 32.24 $ & $ 221.07 \pm 20.81 $ & $ 304.38 \pm 136.95 $ & $ 348.68 \pm 241.32 $ \\
        & \textsc{LODL (Best)}  & $ 365.26 \pm 37.76 $ & $ 474.75 \pm 35.01 $ & $ 623.06 \pm 65.38 $ & $ 1034.88 \pm 46.12 $ \\
        & \textsc{EGL (Best)}  & $ \mathbf{63.01 \pm 4.99} $ & $ \mathbf{72.31 \pm 7.40} $ & $ \mathbf{82.27 \pm 3.89} $ & $ \mathbf{118.99 \pm 2.16} $ \\
        & \textsc{SPO}  & $ - $ & $ 180.51 \pm 30.30 $ & $ - $ & $ - $ \\
        
    \bottomrule
    \end{tabular}

\end{table*} 

%\clearpage

\section{Sensitivity analysis} 
\label{appendix:sensitivity analysis}

In \cref{table:sensitivity} we analyze how $\beta$ influences results. As expected, higher values lead to an increase in the surrogate loss exploitation. However, we observe a counter-intuitive behavior in average regrets, which improve even if less confident estimations take the place of real regrets. We believe these results my be determined by spurious local optima introduced by the GP loss approximation.

\begin{table*}[t]
\centering
\caption{Average regret and average calls per instance for \ouracronym{} with different $\beta$ values on KP-50 with uncertain weights, values and capacity and WSMC-10 with 50 sets.}
\label{table:sensitivity}

\small  
    \begin{tabular}{clcccc}
    
    \toprule
    
    & \textit{Method} & \textit{KP-50 weights} & \textit{KP-50 values} & \textit{KP-50 capacity} & \textit{WSMC-10-50} \\

    \midrule

    % \multicolumn{5}{c}{Average regret} \\
    
    % \midrule

        \multirow{5}{*}{\rotatebox{90}{\textit{Regret}}}
        & $\beta=0.01$  & $ 1.07 \pm 0.14 $ & $  314.02 \pm 263.90 $ & $  3.76 \pm 0.63 $ & $  5.69 \pm 2.17 $ \\
        & $\beta=0.05$  & $ 1.03 \pm 0.16 $ & $  474.31 \pm 136.58 $ & $  3.63 \pm 0.32 $ & $  4.05 \pm 1.86 $ \\
        & $\beta=0.1$  & $ \mathbf{1.01 \pm 0.15} $ & $  176.30 \pm 37.33 $ & $  3.61 \pm 0.33 $ & $  3.87 \pm 1.17 $ \\
        & $\beta=0.5$  & $ 1.06 \pm 0.12 $ & $  \mathbf{74.85 \pm 4.82} $ & $  3.57 \pm 0.35 $ & $  1.77 \pm 0.43 $ \\
        & $\beta=1.0$  & $ 1.04 \pm 0.05 $ & $  75.07 \pm 5.20 $ & $  \mathbf{3.45 \pm 0.31} $ & $  \mathbf{1.18 \pm 0.09} $ \\

    \midrule

    %\multicolumn{5}{c}{Average calls} \\

    %\midrule

        \multirow{6}{*}{\rotatebox{90}{\textit{Slv. calls}}}
        & $\beta=0.01$  & $ 279.16 \pm 12.40 $ & $  148.95 \pm 41.43 $ & $  81.67 \pm 21.51 $ & $  53.65 \pm 7.54 $ \\
        & $\beta=0.05$  & $ 156.53 \pm 23.56 $ & $  57.25 \pm 3.63 $ & $  56.84 \pm 2.95 $ & $  71.68 \pm 43.56 $ \\
        & $\beta=0.1$  & $ 116.74 \pm 13.90 $ & $  55.52 \pm 2.34 $ & $  41.35 \pm 0.84 $ & $  47.09 \pm 2.71 $ \\
        & $\beta=0.5$  & $ 57.75 \pm 2.55 $ & $  \mathbf{40.29 \pm 0.30} $ & $  36.52 \pm 0.13 $ & $  39.73 \pm 1.74 $ \\
        & $\beta=1.0$  & $ \mathbf{41.33 \pm 1.22} $ & $  42.32 \pm 0.09 $ & $  \mathbf{30.04 \pm 0.05} $ & $  \mathbf{31.06 \pm 0.35} $ \\
        
    \bottomrule
    \end{tabular}

\end{table*} 

\end{document}